\documentclass{article}
\usepackage{ijcai22}

\usepackage{times}
\usepackage{soul}
\usepackage{url}
\usepackage[hidelinks]{hyperref}
\usepackage[utf8]{inputenc}
\usepackage[small]{caption}
\usepackage{graphicx}
\usepackage{amsmath}
\usepackage{amsthm}
\usepackage{booktabs}
\usepackage{algorithm}
\usepackage{algorithmic}
\usepackage{amsfonts,amssymb}
\usepackage{color}
\usepackage{multirow}
\usepackage{adjustbox}

\newtheorem{theorem}{Theorem}

\newtheorem{lemma}{Lemma}

\title{Solving the Spike Feature Information Vanishing Problem in Spiking Deep Q Network with Potential Based Normalization}

\author{
	Yinqian Sun$^{1,2}$\and
	Yi Zeng$^{1,2,3,4,5}$\footnote{Contact Author}\and
	Yang Li$^{1,3}$ \\
	\affiliations
	$^1$Research Center for Brain-inspired Intelligence, Institute of Automation, Chinese Academy of Sciences, Beijing, China \\
	$^2$School of Future Technology, University of Chinese Academy of Sciences, Beijing, China\\
	$^3$School of Artificial Intelligence, University of Chinese Academy of Sciences, Beijing , Chin\\
	$^4$National Laboratory of Pattern Recognition, Institute of Automation, Chinese Academy of Sciences, Beijing, China\\
	$^5$Center for Excellence in Brain Science and Intelligence Technology, Chinese Academy of Sciences, Shanghai, China
	\emails
	sunyinqian2018@ia.ac.cn,
	yi.zeng@ia.ac.cn,
	liyang2019@ia.ac.cn
}

\begin{document}

\maketitle

\begin{abstract}


Brain inspired spiking neural networks (SNNs) have been successfully applied to many pattern recognition domains. The SNNs based deep structure have achieved considerable results in perceptual tasks, such as image classification, target detection. However, the application of deep SNNs in reinforcement learning (RL) tasks is still a problem to be explored. Although there have been previous studies on the combination of SNNs and RL, most of them focus on robotic control problems with shallow networks or using ANN-SNN conversion method to implement spiking deep Q Network (SDQN). In this work, we mathematically analyzed the problem of the disappearance of spiking signal features in SDQN and proposed a potential based layer normalization(pbLN) method to directly train spiking deep Q networks. Experiment shows that compared with state-of-art ANN-SNN conversion method and other SDQN works, the proposed pbLN spiking deep Q networks (PL-SDQN) achieved better performance on Atari game tasks.
\end{abstract}

\section{Introduction}

\begin{figure}[tb]
	\centering
	\includegraphics[width=8cm]{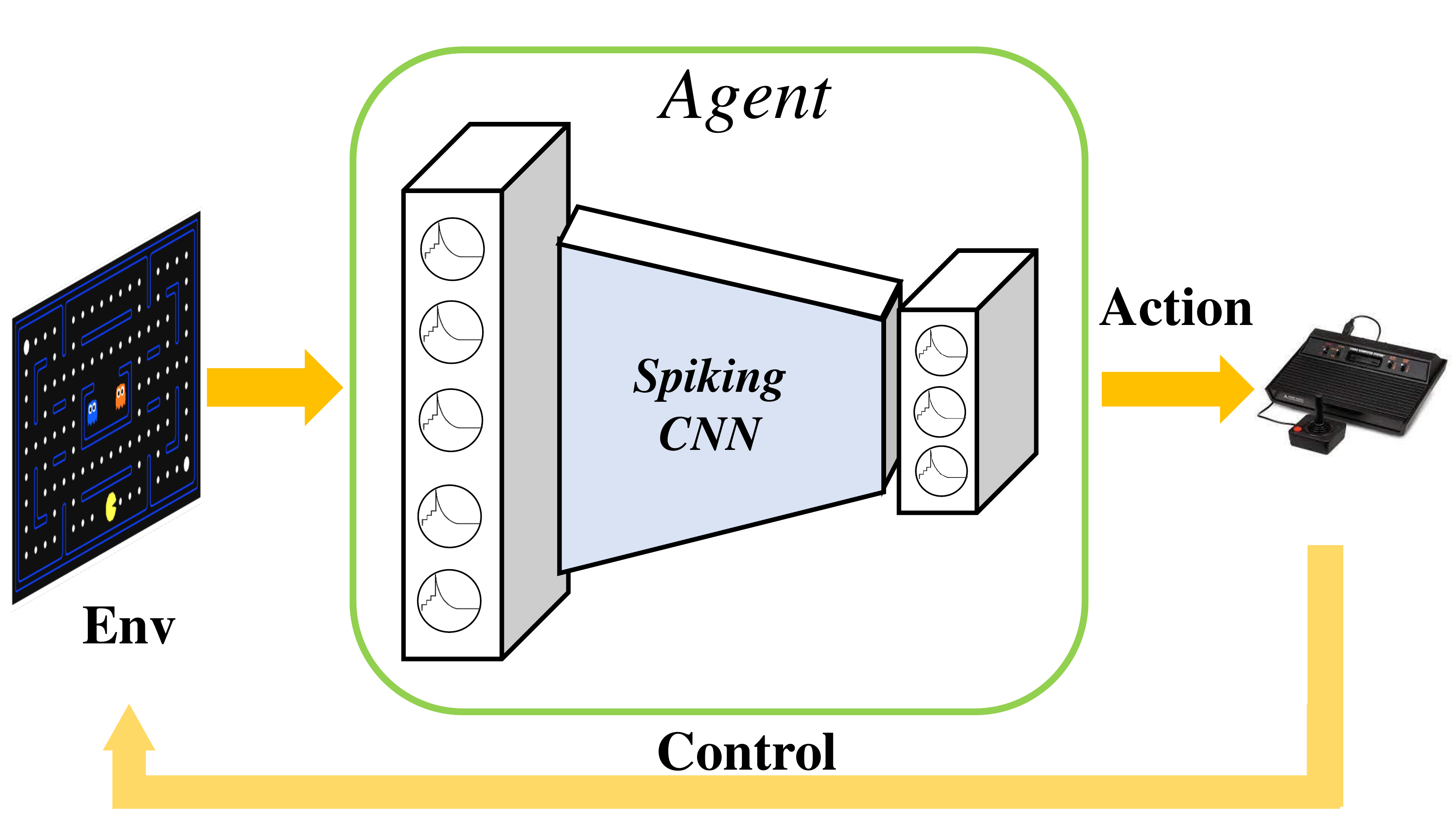}
	\caption{Illusion of spiking DQN.}
	\label{fig:spiking_dqn}
\end{figure}

Inspired by biological brain neurons, the spiking neural network uses differential dynamics equations and spike information encoding methods to build computing node model in neural networks~\cite{maass1997networks}. The traditional artificial neuron models, such as Perceptron and Sigmoids, sum the inputs and then pass it through a non-linear activation function as model output. Different with ANNs, the spiking neurons accept signals from pre-synaptic inputs with special synapses model, and then integrates the post-synaptic potential, firing a spike when the somatic potential exceeds a threshold. The neuron potential is reset when the spike is release to prepare next integrate-and-fire process. According to the complex structure and dynamic characteristics of biological neurons, the spiking neuron model has many forms, including the leaky integrated-and-fire (LIF) model, Izhikevich model, Hodgkin-Huxley model and spike response model. 

Spike neural networks can be applied to different domains of pattern information processing. SNNs achieve competitive performance on many tasks compared with ANN. Spiking Resnet are trained for image classification ~\cite{fang2021deep}, and Spiking-YOLO ~\cite{kim2020spiking} using ANN-SNN conversion method to implement faster and efficient object detection. Brain visual pathway-inspired spiking neural network can deal images features with biologically plausible ~\cite{hopkins2018spiking}. CRSNN ~\cite{fang2021brain} implemented causal reasoning with SNN and Spike-Timing-Dependent Plasticity (STDP).  QS-SNN~\cite{sun2021quantum} takes advantages of spatiotemporal property of spike trans, and process complement information with spiking rate and phase encoding. And ~\cite{zhao2018brain,cox2019striatal} implement basal ganglia-based SNNs models in many decision making tasks. Besides the neuromorphic hardware based on SNNs, likes TrueNorth~\cite{merolla2014million}, SpiNNakers~\cite{furber2014spinnaker} and Loihi~\cite{davies2018loihi} reduces the energy consumption thousands of time than chips based on traditional computing architecture. 

Although there have been previous studies on the combination of SNNs and RL, most of them focus on robotic control problems with shallow networks and few neurons. Reward-modulated  spike-timing-dependent plasticity (R-STDP) is used for training SNN to  control robot keeping within lane. Lele et al proposed SNN central pattern generators(CPG) and leaned with stochastic reinforcement based STDP to control hexapods walking~\cite{lele2020learning}. PopSAN~\cite{tang2020deep} trained spiking actor with a deep critic network and validated on OpenAI gym continuous control benchmarks and autonomous robots. This work adopts actor-critic architecture and explores the combination of deep reinforcement learning and spiking neural network, but they only achieved implementing actor network with SNN, the critic for state-action value estimation was still using ANN. Therefore, they did not totally exploit the low-power advantage of implementing SNN. Because of the optimizing hardness and learning latency, agent based on SNN is difficult to be trained in reinforcement learning tasks. To avoid this difficulty, ANNs to SNNs conversion method~\cite{rueckauer2017conversion} is used to implement DQN with spiking neural network~\cite{patel2019improved,tan2020strategy}. They first trained ANN based DQN policy and then transferred the network weight to SNN, realizing using SNN to plat Atari game.  Zhang et al used knowledge distillation to train student SNN with a deep Q network teacher~\cite{zhang2021distilling}, but the student SNN does not being trained by RL method with reward and not interact with environment. 

Direct training of SNNs can obtain better performance advantages compared with ANNs to SNNs method, and also improve energy efficiency~\cite{wu2019direct,zheng2020going}. Although there are many successful cases of implementing directly trained SNNs model in computer vision tasks, the direct training of SNN in deep reinforcement learning (DRL) model is facing more hardness. One of the most import factors hindering the application of SNN in DRL is the  disappearance of spike firing activity in deep spiking convolutional neural networks. The works in ~\cite{liu2021human,chen2022deep} proposed directly training methods for spiking deep Q networks in RL task, but they do not deal with spiking activity vanishing problem in SDQN. In this work, we mathematically analyzed the problem of the disappearance of spiking signal features in SDQN and proposed a potential based layer normalization(pbLN) method to directly train spiking deep Q networks. Experiments shows our work achieve better performance than SDQN based on ANNs to SNNs method and other trained spiking DQN models.

We summary our contributions as follows:
\begin{itemize}
	\item We analyze how spiking process influence information feature extraction in deep SNNs and found that the binary property of spike hugely dissipates the variance and shift the mean of network inputs. The pattern features of  information are quickly vanishing in spiking deep Q networks. 
	\item We proposes the potential based layer normalization method to keep the unique sensitivity of spiking neuron in deep Q networks.
	\item We construct spiking deep Q network and implement it in gym Atari environments. The spiking deep Q network is directly trained with surrogate function and the experiments show that the pbLN improves the performance of SNNs in RL tasks. 
\end{itemize}

\section{Methods}
In this section, we introduce our work with three aspects. Firstly, we construct spiking deep Q network to estimation state-action value. Secondly we analyze the feature vanishing in SNN and its influences in reinforcement learning. Thirdly, we propose the potential based layer normalization method and train the spiking deep Q network with BPTT algorithm.

\subsection{Spiking Deep Q Network}
\begin{figure*}[tb]
	\centering
	\includegraphics[width=16cm]{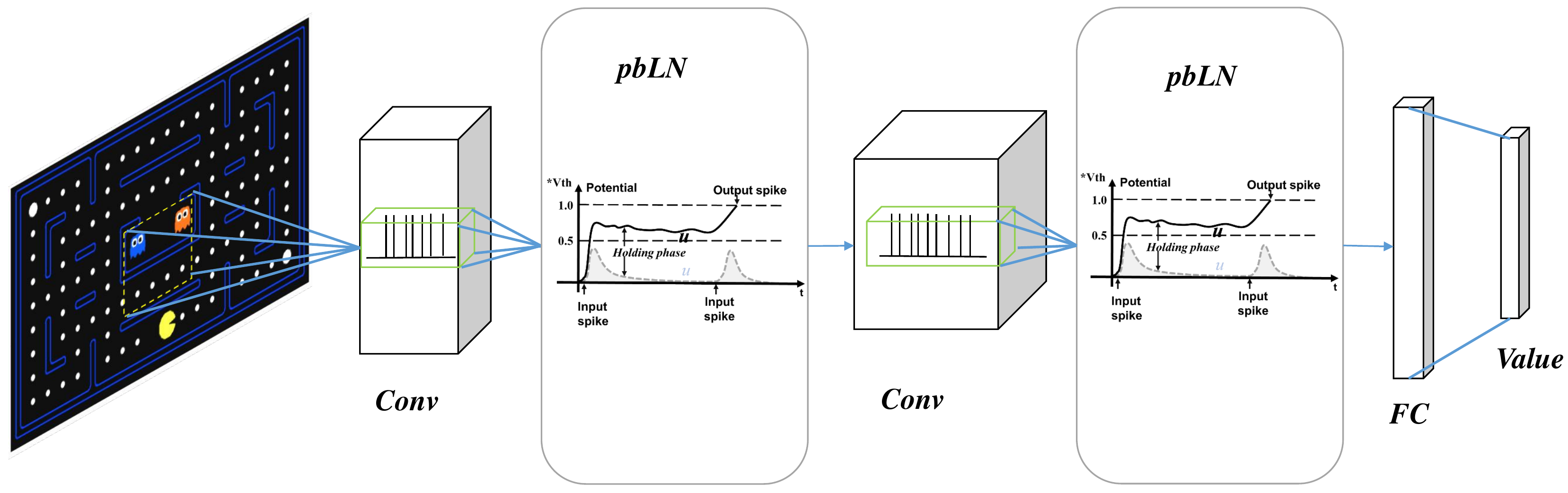}
	\caption{Spiking deep Q network with potential based layer normalization. It has the same network structure as DQN, three layers of convolution and two layers of full connection. The network outputs an estimate of the state-action value, which is used for selection of actions and TD based leaning.}
	\label{fig:sdqn}
\end{figure*}

In order to better reflect the characteristics of SNNs in the reinforcement learning environments, we construct our spiking Q networks as same as the DQN architecture shown in Figure \ref{fig:sdqn}. Raw game screen images get from gym Atari simulation are processed by 3-layers spiking convolution neural networks to generated vision embedding. Then the vision embedding spike trains input to fully connected (FC) spiking neurons population. To generate continuous real value from discrete spikes, we use weighted spike integration to estimate state-action values.


\begin{figure}[tb]
	\centering
	\includegraphics[width=8cm]{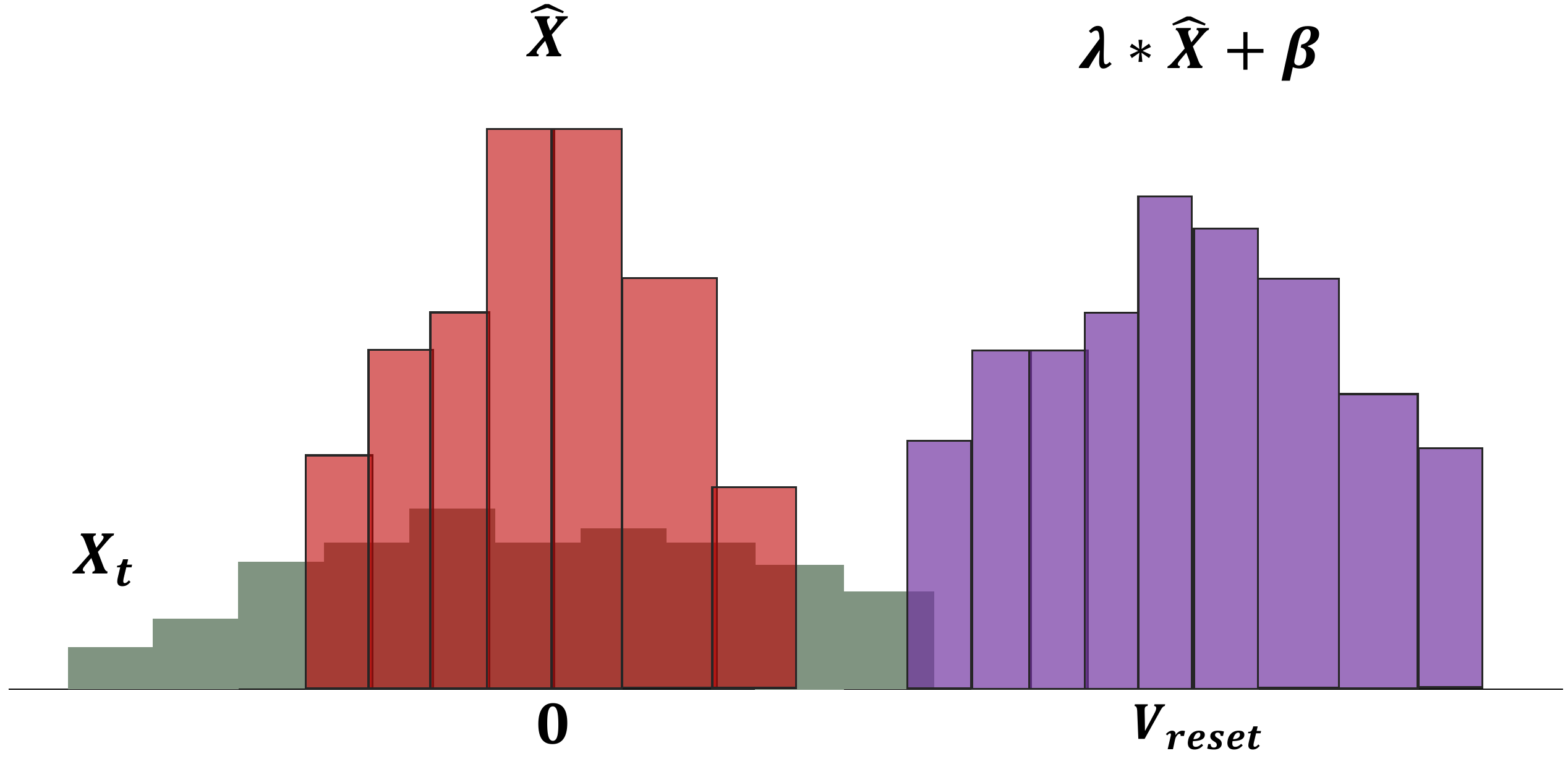}
	\caption{Operations of neural potential based layer normalization.}
	\label{Fig:PotentialShift}
\end{figure}

\begin{figure}[tb]
	\centering
	\includegraphics[width=8cm]{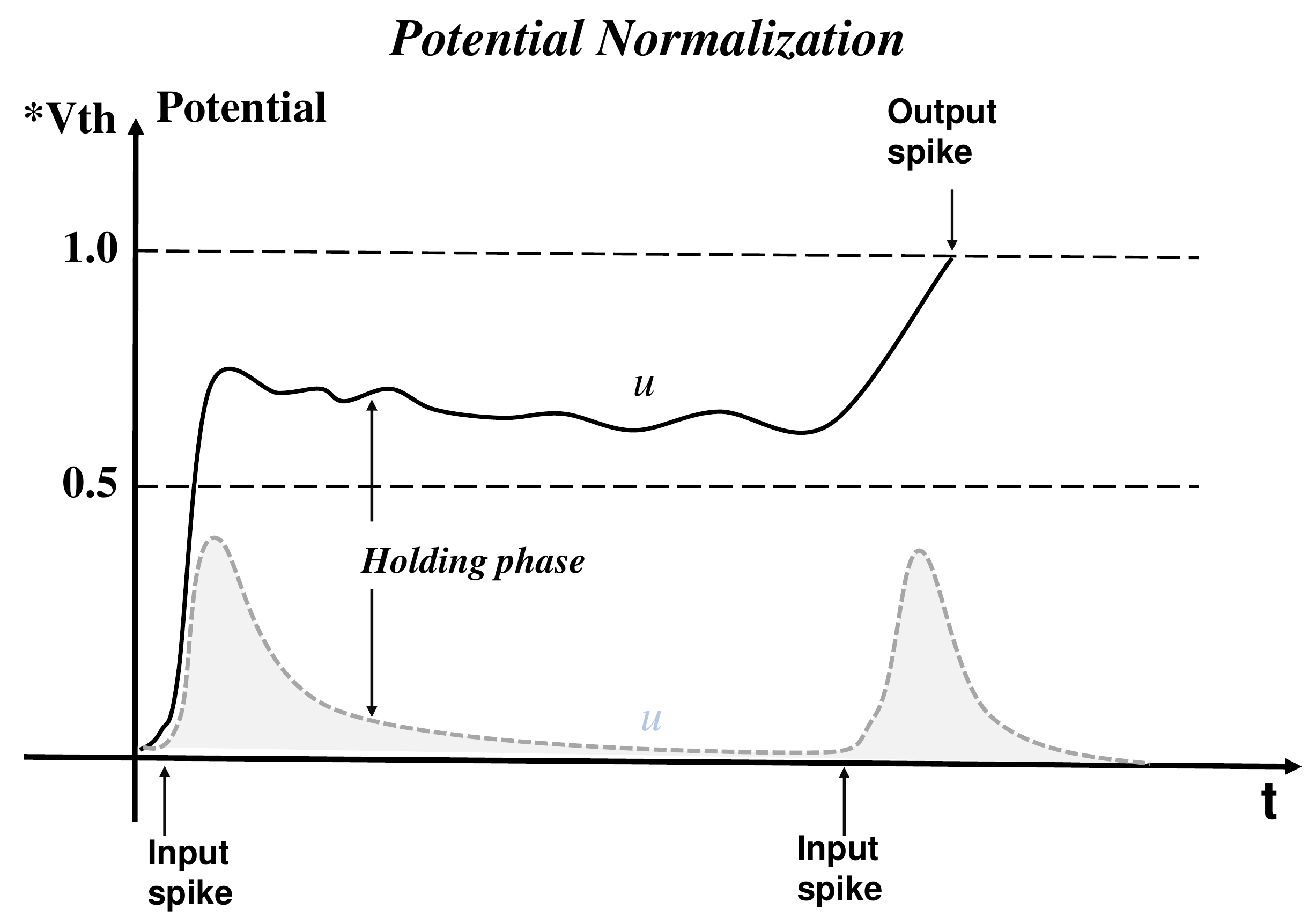}
	\caption{Neuron potential is maintained by normalization method. The gray dotted line is the change of membrane potential of neurons without normalization when receiving spike inputs. And the black solid line is neuron potential changed by normalization. Neurons are difficult to fire when the time interval of external stimulation is relatively large. Instead with normalization operations, the membrane potential is affected by the neighbor neurons and the leakage trend will slow down, which increases the probability of neurons firing spikes.}
	\label{Fig:PotentialNorm}
\end{figure}

Neuron model in the spiking Q Network is adapted from leaky integrate-and-fire (LIF) model. 

\begin{equation}
	\tau\frac{du}{dt} = -u + x
\end{equation}

where $u$ is membrane potential, $\tau$ is decay constant. $x$ is postsynaptic potential (PSP). When the membrane potential exceed thresholds $V_{th}$, the neuron fires a spike and the membrane potential is reset to $V_{reset}$. Here we use Heaviside step function $H$ to implement spiking procedure, and the main neuron dynamic process of our model can be described as follows:


\begin{equation}
	u_{t+1}^l= u_{t}^l + \frac{1}{\tau}(-u_t^l + x_t^l)
	\label{mem_update}
\end{equation}

\begin{equation}
	o_{t+1}^l=H(u_{t+1}^l-V_{th})
	\label{spike_out}
\end{equation}

\begin{equation}
	u_{t}^l=V_{reset}, \quad if \ o_t^l = 1.
	\label{reset_u}
\end{equation}

Let $\alpha=1-\frac{1}{\tau}$ as decay factor of neuron potential, we get the clear format of Eq. \ref{mem_update} as $u_{t+1}^l= \alpha u_{t}^l + (1-\alpha)x_t^l$. The $o_t^l$ represents the number $l$ layer spike outputs. Neuron input $x_t$ of convolutional layers in vision processing is 2D convolution operation on previous layer's spike features and can be written as: 

\begin{align}
	x_{t}^l=& (w^l\ast o_t^{l-1})[m, n] = \sum_{j}\sum_{k}w^l[j,k]o^{l-1}[\nonumber \\ 
	& m-j, n-k]
	\label{Eq:conv}
\end{align}
where $w^l$ is kernel weight of $l$ convolutional layer. In the FC layer, $x_t$ is weighted sum of previous layer spikes:
\begin{equation}
 	x_{t}^l= \sum_{j}w_j^lo_{t,j}^{l-1}
 	\label{inputs}
\end{equation}

State-action value is generated from the time-window mean value of weighted sum spikes output from FC layer as Eq. \ref{eq:Q-gen} and $T$ is the time-windows length of SNN simulation.
\begin{equation}
	q_i = \frac{1}{T}\sum_{t=0}^{T-1}W_i\cdot O_t
	\label{eq:Q-gen}
\end{equation}

Unlike methods using ANN-SNN transformation or DNN-assisted hyper training, our proposed model is directly optimized using the TD error about the network output with target values shown as below:

 \begin{equation}
 	L(w)= \mathbb{E}[(r+\gamma \mathop{max}\limits_{a}Q(s', a',w)-Q(s, a, w))^2]
 	\label{eq:td_error}
 \end{equation}

The proposed deep spiking Q network is directly trained by spatio-temporal backpropagation (STBP) algorithm~\cite{Wu2018Spatio}. For the final output weight $W_i$, we have the derivatives: 
\begin{equation} 
	\frac{\partial L}{W_i} = \frac{\partial L}{\partial Q} \frac{Q}{\partial W_i} = \frac{1}{T}\sum_{t=0}^{T-1}\delta^L O_t
	\label{EQ:derweight}	
\end{equation}
where  $\delta^L$ is Q learning temporal difference
\begin{equation}
	\delta^L = \frac{1}{2}[r+\gamma \mathop{max}\limits_{a}Q(s', a',w)-Q(s, a, w)]
\end{equation}

To optimize the networks weight $w^l_j$, the key temporal chain of derivative is written as:
\begin{align}
	\frac{\partial{L}}{\partial w^l_j} &= \sum^{T-1}_{t=0}\frac{\partial L}{\partial o_{t+1}^l}\frac{\partial o_{t+1}^l}{\partial u_{t+1}^l}\frac{\partial u_{t+1}^l}{\partial w_j^l}  \\
					& = \sum_{t=0}^{T-2}\frac{\partial L}{\partial o_{t+1}^l}\frac{\partial o_{t+1}^l}{\partial u_{t+1}^l} [\sum_j(1-\alpha)o_{t,j}^{l-1}+\alpha \frac{\partial u_t^l}{\partial w_j}] \\
\end{align}

At the non-differentiable point of neuron firing a spike, we use surrogate function to approximate the derivative of $o_t$ with respect to $u_t$  as follows:


\begin{equation}
	\frac{\partial o_{t+1}^l}{\partial u_{t+1}^l} = \frac{2\tau}{4+(\pi \tau u_{t+1}^l)^2}
\end{equation}

\begin{algorithm}[tb]
	\caption{Proceeding process of proposed SNN, to simplify the analysis supposing all layers are fully connected.}
	\label{alg:Proc_SNN}
	\textbf{Input}: Observation state $S$ as $x_0^0$ \\
	\textbf{Parameter}: Decay factor $\tau$; simulation time-window $T$; membrane potential threshold $V_{th}$; layer numbers $N$. \\
	\textbf{Output}: State-action values $Qs$. \\
	\begin{algorithmic}[1] 
		\STATE Initialize neuron weight $w^l$ $\forall \ l=0,1,2,...,N-1$. \\
		\FOR{$t=0$ to $T-1$}
		\FOR{$l=0$ to $N-1$}
		\STATE Calculate layer neuron inputs $x_t^l \leftarrow \sum_j^lw_j^lo_{t, j}^l$. \\ 
		\STATE Membrane potential update $u_{t+1}^l \leftarrow \alpha u_l^t + (1-\alpha)*pbLN(x_t^l)$. \\
		\STATE Layer spike outputs $o_{t+1}^l \leftarrow H(U_{t+1}^l-V_{th})$. \\
		\ENDFOR
		\ENDFOR 
		\STATE Calculate state-action values $Q \leftarrow \frac{1}{T}\sum_{t=0}^{T-1}W_i O^{N-1}_{t+1}$. \\
		\STATE \textbf{return} $Q$
	\end{algorithmic}
\end{algorithm}

\subsection{The feature information vanishing in spiking neural networks}

For $N$ layers SNN the proceeding procedure can be written as Algorithm~\ref{alg:Proc_SNN}. Because of spiking neuron model needs accumulate the membrane potential in whole simulation time window, the information is processed along both time and space dimensions. Considering the binary distribution of spikes $o_t$, the means of spike $\mathbb{E}(o_t)$ and variance $\mathbb{D}(o_t)$ have below property $\mathbb{E}[(o_t)^2]=\mathbb{E}[o_t]$ and $\mathbb{D}(o_t)=\mathbb{E}(o_t)(1-\mathbb{E}(o_t))$. 

\begin{lemma}
	
	Let $\psi(i, j)=(1-\alpha)^2\alpha^{2(j-i)}$, synapse weight $W$ is randomly initialized with $\mathbb{E}(W)=0$ and  is independent with $o_t$, the variance of neuron potential can be get as:
	\begin{equation}
		\mathbb{D}(u_{t+1}^l) = \mathbb{D}(W^l)\sum_{i=0}^t\psi(i, t) \mathbb{E}[o_i^{l-1}] 
		\label{EQ:poten}
	\end{equation}
	\label{LEMMA:poten}
\end{lemma}


Supposing the neuron in layer $l$ fired the first spike after $t$ time steps stimulation signal input, we denote the variance of neuron membrane potential at time $t+1$ and layer $l$ as Eq. \ref{EQ:poten}. Lemma~\ref{LEMMA:poten} shows  the relationships of neuron membrane potential $u_t$ with previous inputs spike history and synaptic weight. And the presynaptic spike inputs' effects decays with time factor $\alpha$,  with $\psi(i, j) \in (0, 1]$. See Proofs section for details.





\begin{theorem}
	Let $\varepsilon=\frac{\mathbb{D}(W^l)}{2V_{th}^2}$, neuron in layer $l$ firing spikes with:
	\begin{equation}
		\mathbb{E}(o_{t+1}^l) \le \varepsilon \mathbb{E}[\sum_{i=1}^t o_{i}^{l-1}]	
		\label{EQ:spike_expe}	
	\end{equation}
	\label{Theo:spike_expe}
\end{theorem}


%
%

$\varepsilon$ is the loss ratio of signal transmitted by spiking neural networks. And if $\varepsilon < 1$, the mean of neuron's spikes $\mathbb{E}(o_t)$ are tends to zero with the increase of the number of layers, and the variance of spikes $\mathbb{D}(o_t)$ is also decreasing, which results neuronal firing spikes vanish rapidly and the deep layer of SNNs are very prone to no spiking activity. This problem makes deep SNNs lose signal features in information processing. And in deep spiking convolutional neural networks, which use the locally connections and sharing weight operations, the spike signal disappearance problem is more obvious, which makes it hard for the deep SCNN to be directly trained, and weakens the performance of SNN models.


\subsection{Potential Based Layer Normalization}
According Eq.\ref{EQ:spike_expe}, the problem of spike  information vanishing in SNNs can be alleviated initializing synapse weights with greater distributional variance or set the spiking neuron model with little potential threshold. But increasing $\mathbb{D}(W)$ will damage performance and make the model difficult to converge. Besides, too small threshold potential will make neurons too active to distinguish effective information. To solve this contradiction, some works using the Potential normalization methods in spiking neural networks, such as NeuNorm~\cite{wu2019direct} uses auxiliary neuron to add spikes together and proposes inputs with scalar norm, and tdNorm~\cite{zheng2020going} extends batch normalization to time dimensions.

But these methods are suitable for supervised learning tasks such as image classification or object detection, because in those task, the SNNs are trained with batch data inputs. Different from supervised learning, the environment information of reinforcement learning is more complex. Firstly, in RL tasks spike vanishing problem of deep SNN models is quite serious. For example, we counted the spiking deep Q network firing activity distribution of each layer when it was applied to Atari games. The statistic evidence shows that the SDQN suffers spiking information reduction in deep layers. Secondly, different with supervised learning, SNN agents in RL have no invariant and accurate learning labels, and need to interact with the environment to collect data and reward information. The hysteresis of learning samples makes the SDQN model unable to effectively overcome the drawbacks caused by the disappearance of spike signal in output layers. Thirdly, the inputs format in RL task are not batched, so the normalization methods used in supervised learning can not be applied to SDQN.

In this work, we propose a potential based layer normalization method to solve the spike activity vanishing problem in SDQN. We apply the normalization operation methods on PSP $x_t$ in convolution layers. The previous layers' spikes are processed as Eq.~\ref{Eq:conv}, and further normalized as follow:
\begin{equation}
	\hat{x_t} = \frac{x_t-\bar{x}_t}{\sqrt{\sigma_{x_t}+\epsilon}}
	\label{Eq:norm}
\end{equation}	
	
\begin{equation}
	\bar{x}_t = \frac{1}{H} \sum_{i=1}^Hx_{t, i}
\end{equation}	

\begin{equation}
	\sigma_{x_t} = \frac{1}{H}\sqrt{\sum_{i=1}^H(x_{t, i}-\bar{x}_t)}
\end{equation}
where in convolution layer $H=C\times H \times W$ with $C$ for channels number, and $[H,W]$ is features shape in each channel. 

PSP $x_t$ is normalized into a distribution with zero mean and one variance shown as Eq.~\ref{Eq:norm}. This normalization method is different with NeuNorm and tdNorm. NeuNorm applied normalization operation on neuron spikes $o_t$. And tdNorm used batch normalization method on time dimension, which needs to calculated $[x_{t+1}, x_{t+2},...,x_{t+T}]$ in advance. 

Normalizing $x_t$ will weaken the characteristic information contained in the feature maps, and the zero means and one variance does not suit to spiking neuron. Thus we adapted the LIF neuron model to:

\begin{equation}
	u_{t+1} = \alpha u_t + (1-\alpha)[\lambda_t*\hat{x}_t + \beta_t]
\end{equation}

\begin{equation}
	\lambda_0 = V_{th}-V_{reset}, \quad \beta_0=V_{reset}
	\label{Eq:init}
\end{equation}
where $\lambda_t$ and $\beta_t$ are learnable parameters, which are initialized at beginning as Eq.~\ref{Eq:init}. The process of pbLN changes the distribution of neural PSP and depicted as Figure ~\ref{Fig:PotentialShift}. The parameter $\lambda_t$ has the same effect of increasing $\mathbb{D}(W)$, and $\beta_t$ plays a role of dynamic firing threshold. By separating the learnable parameters, the SNN model avoids the oscillation of the learning process caused by increasing $\mathbb{D}(W)$ and the over-discharge of neurons caused by the threshold being too small, which reduces the information processing capability of the model. 

The effect of pbLN on membrane potential is shown in Figure \ref{Fig:PotentialNorm}. When the spike signal of the previous layer is input, the membrane potential begins to rise. Comparing with LIF model, neurons with function of pbLN are affected by neighbors and can holding the  membrane potential values so that the neuron can firing a spike as long as it receives little input in the future. This puts the neuron in a easy-to-fire state where it can process long time interval signal and reduces the loss of features when passing on the input information.

\section{Proofs}

\begin{proof}
	
	Lemma \ref{LEMMA:poten}
	
	In this work the synapse weight value is sampled from uniform distribution $U(-k, k)$, $k$ is constant and related with inputs signal dimension.
	We get the variance of postsynaptic potential $x_t$ as
	0
	\begin{align}
		\mathbb{D}(x_t^l) &= \mathbb{D}(W^lo_t^{l-1}) \nonumber \\
		&= \mathbb{E}[(W^l)^2(o_t^{l-1})^2] - \mathbb{E}[(W^l)]\mathbb{E}[(o_t^{l-1})]  \nonumber \\
		& = \mathbb{E}[(W^l)^2]\mathbb{E}[(o_t^{l-1})^2] \nonumber \\
		& = [\mathbb{D}(W^l) + \mathbb{E}^2(W^l)]\mathbb{E}(o_t^{l-1})  \nonumber \\
		& = \mathbb{D}(W^l)\mathbb{E}(o_t^{l-1})
	\end{align}
	Then, we use the recursive method to deduce the relationship between the membrane potential $u_t$ variance and the presynaptic spikes:
	\begin{align}
		\mathbb{D}(u_{t+1}^l) &= \mathbb{D}(\alpha u_t^l + (1-\alpha)x_t^l) \nonumber \\
		& = \alpha^2\mathbb{D}[(u_t^l)^2] + (1-\alpha)^2\mathbb{D}[x_t^l] \nonumber \\
		&=  \alpha^2\mathbb{D}[(u_t^l)^2] + (1-\alpha)^2\mathbb{D}(W^l)\mathbb{E}(o_t^{l-1}) \nonumber \\
		& = \mathbb{D}(W^l)\sum_{i=0}^t\mathbb{E}[o_i^{l-1}]\alpha^{2(t-i)}(1-\alpha)^2
	\end{align}
	
\end{proof}

%
%
%
%

\begin{proof}
	
	Theorem \ref{Theo:spike_expe}
	
	The membrane potential $u_{t+1}$ accumulates  the former $t$ times input spikes $o_t$. And the synapse weight is initialized $\mathbb{E}(W^l)=0$ 
	
	\begin{align}
		\mathbb{E}(u_{t+1}^l) &= \alpha\mathbb{E}(u_t^l) + (1-\alpha)\mathbb{E}(x_t^l) \nonumber \\
		&= \sum_{i=0}^t\alpha^{t-i}(1-\alpha)\mathbb{E}(x_i^l) \nonumber \\
		&= \sum_{i=0}^t\alpha^{t-i}(1-\alpha)\mathbb{E}(W^l)\mathbb{E}(o_i^{l}) \nonumber \\
		&= 0
	\end{align}

	According to the Chebyshev's inequality:
	
	\begin{align}
		\mathbb{E}(o^l_{t+1}) &= P(u^l_{t+1}>V_{th}) \nonumber \\
		&= \frac{1}{2} P(|u_{t+1}^l-\mathbb{E}(u_{t+1}^l)|>V_{th}) \nonumber \\
		& \le \frac{\mathbb{D}(u_{t+1}^l)}{2V^2_{th}}
	\end{align}

	In spiking neural model time constant $\tau\ge 1.0$, the range of decay factor $\psi(i, j) \in [0, 1)$ then we get 
	
	\begin{align}
	\mathbb{E}(o_{t+1}^l) & \le \frac{\mathbb{D}(u_{t+1}^l)}{2V^2_{th}} \nonumber \\
						  & \le \frac{\mathbb{D}(W^l)}{2V^2_{th}} \sum_{i=0}^t \psi(i,t)\mathbb{E}[o_i^{l-1}] \nonumber  \\	
						  & \le	\frac{\mathbb{D}(W^l)}{2V^2_{th}}  \mathbb{E}[\sum_{i=0}^t o_i^{l-1}] 
	\end{align}
\end{proof}

\section{Results}
PL-SDQN is a spiking neural network model based on LIF neurons and has the same network structure as traditional DQN. It contains three convolutional layers with "c32k8-c64k4-c64k3" neural structure. The hidden layers is fully connected with 512 neurons, and the output is 10 values as the weighted summation about the hidden layers outputs. We directly trained PL-SDQN on  reinforcement learning task. The results shows spike deep Q networks combined  with potential based layer normalization method can achieves better performance on Atari games than traditional DQN and ANN-to-SNN conversion method.

\subsection{Statistic evidence of spiking activity reduction in deep layers}
We counted each layer's firing spikes of SDQN to show the deep layer spike vanish phenomenon and the promoting effect of pbLN method. The SDQN model is firstly initialized by random synaptic weigh, and then are used to play Atari game. We calculated the ratio of neurons with firing activity to the total number of neurons in each layer.

\begin{figure}[tb]
	\centering
	\includegraphics[width=8cm]{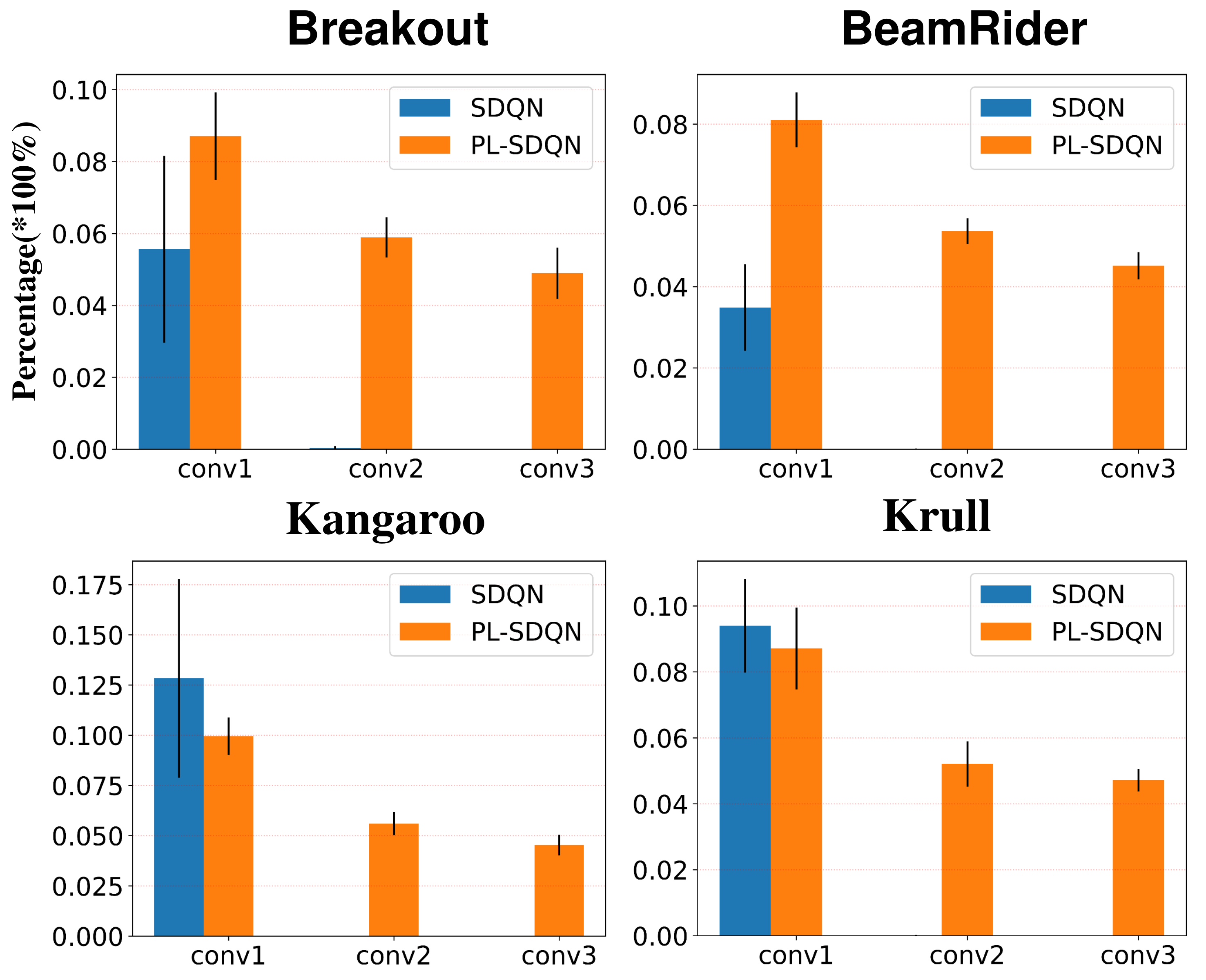}
	\caption{Fire rates on different convolutional layers of SDQN. The blue bars depict the spike fire rate in SDQN without normalization. And the orange bars are for our model PL-SDQN. The black vertical line on top of each bar is standard deviation of 10 experimental data.}
	\label{Fig:fire_record}
\end{figure}

We tested each game ten times, and counted the firing rate of each layer. The average and standard deviation of these experiments are shown in Figure~\ref{Fig:fire_record}. The results shows convolution layers in SDQN is difficult to transfer spiking activities. 
Spike from the first layer (conv1) are rarely transmitted to the next layer. There is almost no spike firing activity in the second (conv2) and third (conv3) layers. 

Comparing with the vanishing problems in SDQN, the proposed pbLN method improves the deep layers spiking activities. The bottom rows in Figure~\ref{Fig:fire_record} row shows that pbLN method does not only increase the activity of the first layer to make later layers fire more spikes, it also improves the inner sensitivity of each layer of the network to spike inputs.

\subsection{Performance and analysis on Atari games}
\begin{figure*}[tb]
	\centering
	\includegraphics[width=18cm]{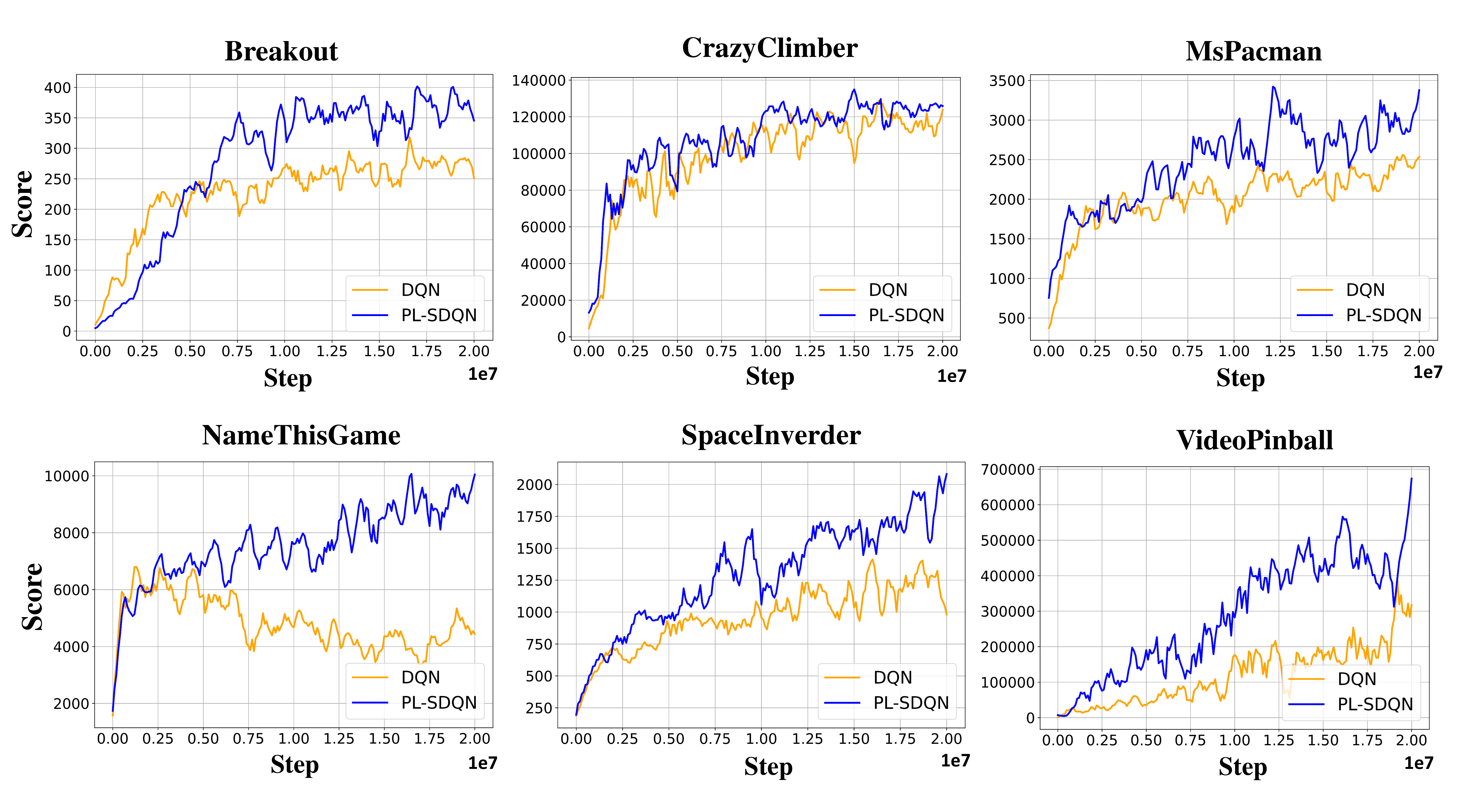}
	\caption{PL-SDQN performance on Atari games.}
	\label{Fig:game_res}
\end{figure*}

We compared our model with vanilla DQN model and ANN-SNN conversion based SDQN model, the performance are got on 16 Atari games.  All models are trained directly with same settings  and optimized by Adam methods as Table~\ref{Tab:parameter}. The ANN-SNN conversion based SDQN are implemented the state-of-art works proposed by ~\cite{li2022efficient} with simulation time window $T_{con}=256$. And the PL-SDQN is directly trained by STBP method proposed before section with simulation times are set as $T=16$.  We train all of the model for 20 million frame steps. We conducted ten rounds of tests and recorded the mean and standard deviation of scores. The results are showed in Figure \ref{Fig:game_res} and Table \ref{Tab:game_res}.

\begin{table}
	\centering
	\begin{tabular}{lll}
			\toprule
			\textbf{Parameters} &  \textbf{Value} & \textbf{Description} \\
			\midrule
			 $\tau$      & 2    & Membrane time constant  \\
			 $V_{th}$    & 1.0  &  Threshold potential \\
			 $V_{reset}$ & 0.0  &  Reset potential \\
			 $T$         & 16   &  Simulation time window  \\ 
			 $\gamma$  & 0.99 &  Discount factor of DQN\\	
			 $T_{con} $ & 256 & Conversion time window\\
			 lr  & 0.0001 &  Learning rate\\	
			\bottomrule
		\end{tabular}
	\caption{Settings of models and experiments.}
	\label{Tab:parameter}
\end{table}

\begin{table*}
	\centering
	\begin{tabular}{l l l l  l l l l l l}	
	 \toprule[2pt]
	& \multicolumn{3}{c}{\textbf{DQN}} & \multicolumn{3}{c}{\textbf{ANN-SNN Conversion}} &\multicolumn{2}{c}{\textbf{PL-SDQN}} \\
	  \textbf{Game} & Score & $\pm$std & (Score$\%$)  & Score & $\pm$std &(Score$\%$)   & Score & $\pm$std & (Score$\%$)   \\
	  \midrule[1pt]
	  Atlantis     & 3049750.0 & 161861.4 & (5.31$\%$)   & 3046920.0 & 1348868.6  &(44.27$\%$) & \textbf{3267760.0} & 86339.7  &(2.64$\%$) \\
	  BeamRider    & 10423.2   & 2245.1   & (21.54$\%$)  & 10449.0   & 2620.4     &(25.08$\%$) & \textbf{11480.4}   & 3342.4   &(29.11$\%$) \\
	  Boxing       & 99.3      & 0.9      & (0.91$\%$)   & 98.6      & 3.2        &(3.25$\%$)  & \textbf{99.5}      & 0.9      &(0.90$\%$) \\
	  Breakout     & 343.1     & 41.3     & (12.04$\%$)  & 352.2     & 64.7       &(18.37$\%$) & \textbf{427.7}     & 140.1    &(32.76$\%$) \\
	  CrazyClimber & 139420.0    & 11530.6  & (8.27$\%$)	 & 128380.0  & 23239.8    &(18.10$\%$) & \textbf{147950.0}    & 30632.1  &(20.70$\%$) \\
	  Gopher       & \textbf{38662.0}     & 28245.6  & (73.06$\%$)	 & 22438.0	 & 10076.7    &(44.91$\%$) & 24064.0   & 12355.3  &(51.24$\%$) \\
	  Jamesbond    & 1445.0    & 1572.0   & (108.79$\%$) & 1420.0    & 190.0      &(13.38$\%$) & \textbf{1460.0}    & 2150.1   &(147.27$\%$) \\
	  Kangaroo	   & 12680.0   & 208.8    & (1.65$\%$)   & 13850.0     & 1132.5   &(8.17$\%$)  & \textbf{14500.0}     & 845.0    &(5.83$\%$) \\
	  Krull	       & 10271.0     & 1365.5   & (13.29$\%$)  & 10923.0     & 513.0        &(4.70$\%$)  & \textbf{11807.0}     & 568.2    &(4.81$\%$) \\
	  MsPacman	   & 2964.0	   & 711.7    & (24.01$\%$)	 & 3691.0	     & 434.8      &(11.78$\%$) & \textbf{4077.0}	   & 1292.5   &(31.70$\%$) \\
	  NameThisGame & 7732.0      & 1289.2   & (16.67$\%$)  & 8115.0    & 1702.1     &(20.97$\%$) & \textbf{12202.0}     & 2210.7   &(18.12$\%$) \\
	  RoadRunner   & 1310.0    & 764.8   & (58.38$\%$)   & 1072.0   & 329.2     &(19.34$\%$)  & \textbf{51930.0}     & 4714.4   &(9.08$\%$)\\
	  SpaceInvaders& 1728.5    & 461.6    & (26.71)      & 1760.0    & 483.5      &(27.47$\%$) & \textbf{2433.5}   & 574.7    &(23.62$\%$) \\
	  StarGunner   & 53050.0	   & 1342.5   & (2.53$\%$)   & 55910.0   & 12796.9    &(22.89$\%$) & \textbf{63560.0}     & 4064.7   &(6.40$\%$) \\
	  Tutankham	   & 262.0       & 28.9     & (11.03$\%$)  & 254.5     & 55.4       &(21.77$\%$) & \textbf{271.5}     & 70.4     &(25.93$\%$) \\
	  VideoPinball & 507442.5  & 327189.1 & (64.48$\%$)  & 552917.6  & 200852.5   &(36.33$\%$) & \textbf{673553.0}    & 10066.2  &(1.49$\%$) \\ 
		\bottomrule[2pt]		
		\end{tabular}
	\caption{Details of Atari game experiments. The vanilla DQN, ANN-SNN conversion based SDQN and our proprosed model PL-SDQN are compared. We test these models 10 rounds and record the mean and standard deviation (std) of raw scores.}
	\label{Tab:game_res}
\end{table*}

PL-SDQN model that we proposed achieves better performance than vanilla DQN and conversion based SDQN model. The scores in Table \ref{Tab:game_res}. The data in the Table \ref{Tab:game_res} shows that our model has achieved performance advantages over the other two methods in 15 games. And the curves on Figure~\ref{Fig:game_res} represent that our method achieves faster and more stable learning than the vanilla DQN.

\subsection{Experiments on potential normalization effects}
In order to show the improvement of our proposed pbLN method on the spiking deep Q model, we compared PL-SDQN with other directly trained SDQN model in  papers ~\cite{liu2021human,chen2022deep}. Because the other works uses different DQN backbones, we recorded the ratio of SDQN model scores and the special DQN methods the authors used in their paper.

\begin{table*}
	\centering
				\begin{tabular}{p{3.0cm} p{3.0cm} p{3.0cm} l }	
					\toprule[2pt]
					&\textbf{SDQN(Liu2021)} & \textbf{SDQN(Chen2022)}  &\textbf{PL-SDQN(Ours)} \\
					\textbf{Game} & (DQN$\%$) & (DQN$\%$) & (DQN$\%$)    \\
					\midrule[1pt]
					Atlantis  &98.79$\%$  & 84.24$\%$ & \textbf{107.15$\%$} \\
					BeamRider &97.48$\%$ &99.57$\%$ & \textbf{110.14$\%$} \\
					Boxing    &  99.17$\%$  & \textbf{298.23$\%$}& 100.20$\%$ \\
					Breakout  & 90.86$\%$ & \textbf{144.38$\%$} & 124.66$\%$ \\
					CrazyClimber &102.82$\%$  & \textbf{109.79$\%$}& 106.12 $\%$ \\
					Gopher    & 95.78$\%$ & \textbf{148.96$\%$}& 62.24 $\%$ \\
					Jamesbond   & \textbf{127.57$\%$} & 113.92$\%$ & 101.04 $\%$\\
					Kangaroo    &\textbf{214.49$\%$} &  94.56$\%$& 114.35 $\%$ \\
					Krull	     &106.77$\%$  &  28.69$\%$ & \textbf{137.55$\%$}\\
					NameThisGame &98.85$\%$ & 152.41$\%$ & \textbf{157.81} $\%$ \\
					RoadRunner   &89.72$\%$ & 917.26$\%$ & \textbf{3964.12$\%$}\\
					SpaceInvaders &80.5$\%$ & 106.85$\%$ &\textbf{140.79$\%$}  \\
					StarGunner  &112.96$\%$  &\textbf{153.73$\%$}& 119.81$\%$\\
					Tutankham	&103.90$\%$  & \textbf{280.48$\%$}& 103.63$\%$\\
					VideoPinball &87.01$\%$ & \textbf{159.98$\%$}& 132.73$\%$ \\ 
					\hline \\
					$total \ge 100 \%$ & 6/15 & 11/15 & \textbf{14/15}\\
					\bottomrule[2pt]		
				\end{tabular}
				\caption{The compare of our PL-SDQN model with state-of-art spiking deep Q networks. To show the SNN advantage, we record the percentage as $SDQN/DQN *100\%$. And the result shows total number of games paper's models get better scores than DQN backbone in corresponding works.}
				\label{Tab:compare}
			\end{table*}
		
The result in Table ~\ref{Tab:compare} shows our PL-SDQN model achieves more stable and better performance than SOTA SDQN methods. Although SNN model in  ~\cite{chen2022deep} has achieved good performance improvement, the PL-SDQN has better generalization and robustness on more experiments for successfully surpassed DQN benchmarks on 14 games out of a total of 15 games tested.

We analyze that the reason why our model has an advantage in the test game is that different game environments have different strengths of the input signal to the SNN. The spiking activity vanishing in deep layers of SNN reduces the performance of the SDQN model. The proposed pbLN method can well counteract the impact of the change of input signal on the performance of the model, so as to better improve the ability of spiking neural networks in reinforcement learning task.

\section{Conclusion}
In this work we directly trained the deep spiking neural networks on Atari game reinforcement learning task. Because of the characteristics of discrete bias and hard to optimization problem, spiking neural network is difficult to be applied to the field of reinforcement learning in complex scenarios. We mathematically analyze the reasons why spiking neural networks are difficult to generate firing activity, and propose a potential based layer normalization method to increase spiking activity i deep layers of SNN. This method can increase the firing rate of the deep spiking neural network, so that the input information features can be saved to the output layer. .And the experiment results shows that compared with vanilla DQN and ANN-SNN conversion based SDQN methods, our PL-SDQN model achieves better task performance. Besides our model has better generalization and robustness compared to other directly trained SDQN method on Atari game reinforcement learning tasks. 

%
%
%
%
%
%

\clearpage

\bibliographystyle{named}
\bibliography{spiking-DQN_Refs}

\end{document}